\documentclass[letterpaper,journal]{IEEEtran}

\usepackage{newtxtext}
\usepackage{newtxmath}

\makeatletter
\def\bfseries{\fontseries\bfdefault\selectfont}
\makeatother
\usepackage{amsmath}

\usepackage{amssymb}

\usepackage{algorithm}
\usepackage{algorithmic}
\usepackage{array}
\usepackage{graphicx}
\usepackage{textcomp}
\usepackage{stfloats}
\usepackage{url}
\usepackage{verbatim}
\usepackage{cite}

\usepackage{booktabs,tabularx}
\usepackage[dvipsnames,table]{xcolor}
\usepackage{colortbl}

\newcolumntype{L}[1]{>{\raggedright\arraybackslash}p{#1}}
\newcolumntype{C}[1]{>{\centering\arraybackslash}p{#1}}
\newcolumntype{Y}{>{\raggedright\arraybackslash}X}

\colorlet{surfA}{gray!05}
\colorlet{embedA}{RoyalBlue!05}
\colorlet{cycleA}{Orange!8}
\colorlet{oursA}{ForestGreen!10}

\colorlet{surfB}{gray!35}
\colorlet{embedB}{RoyalBlue!45}
\colorlet{cycleB}{Orange!55}
\colorlet{oursB}{ForestGreen!60}

\newcommand{\legendbox}[2]{\textcolor{#1}{\rule{6pt}{6pt}}\,#2}

\usepackage{pifont}
\newcommand{\cmark}{\textcolor{green!70!black}{\ding{51}}}
\newcommand{\xmark}{\textcolor{red!80!black}{\ding{55}}}

\definecolor{wacvblue}{rgb}{0.21,0.49,0.74}
\usepackage[pagebackref,breaklinks,colorlinks,allcolors=wacvblue]{hyperref}
\usepackage[capitalize]{cleveref}

\usepackage[caption=false,font=normalsize,labelfont=sf,textfont=sf]{subfig}

\makeatletter
\newsavebox{\sf@box}

\makeatother

\usepackage{capt-of}
\usepackage{xcolor}
\usepackage{subcaption}
\providecommand{\citep}[1]{\cite{#1}}
\providecommand{\citet}[1]{\cite{#1}}
\providecommand{\citeauthor}[1]{\cite{#1}}
\providecommand{\citeyear}[1]{\cite{#1}}

\makeatletter
\AtBeginDocument{%
  \let\@oldmaketitle\@maketitle
  \renewcommand{\@maketitle}{%
    \@oldmaketitle
    \begin{center}
      \setcounter{figure}{0}
      \includegraphics[width=\linewidth]{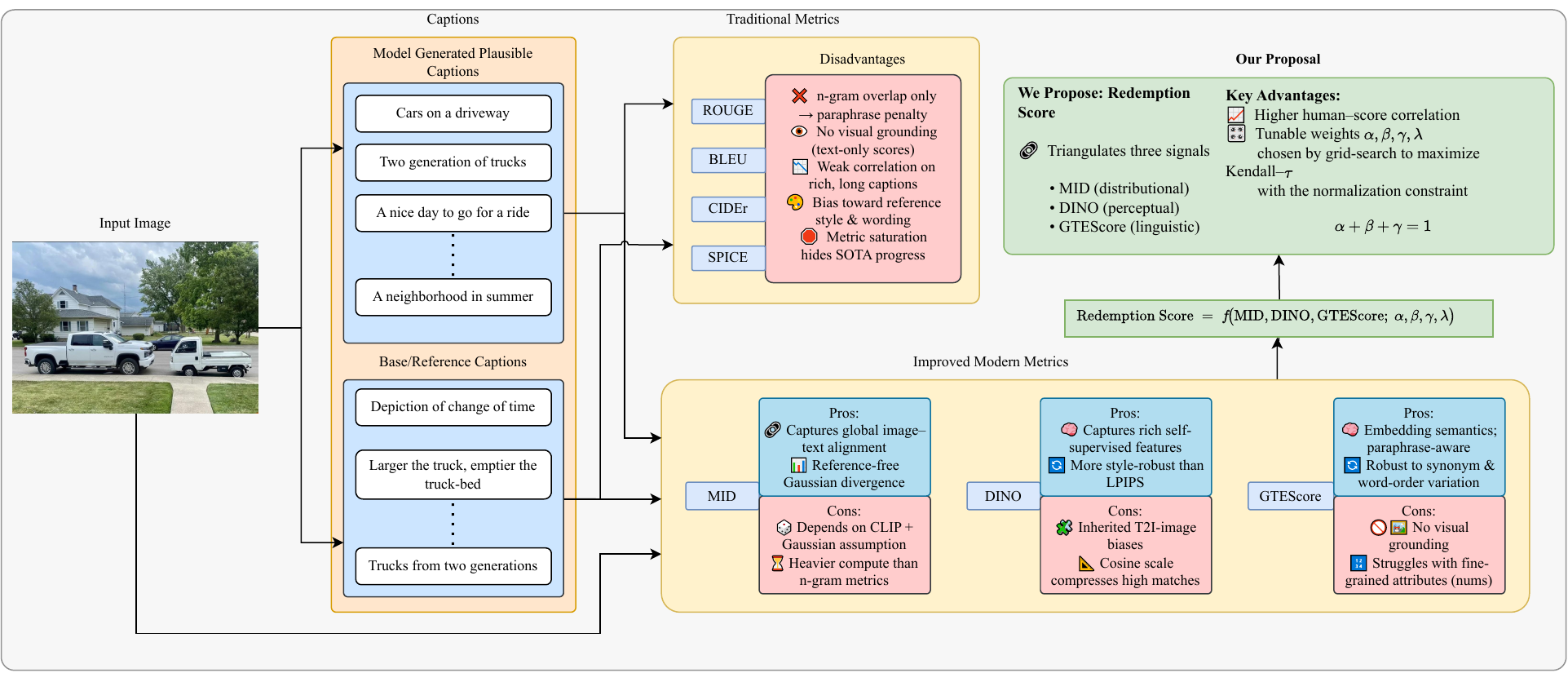}
      \vspace{-4pt}
      \captionof{figure}{Abstract figure depicting the rationale for the study. Left side shows inputs towards various metric systems whereas the proposed method is highlighted in green on the right. Advantages and disadvantages of current metrics are highlighted in light blue and red respectively.}
      \label{fig:abstract_figure}
    \end{center}
    \vspace{-10pt}
  }%
}
\makeatother

\title{Redemption Score: A Multimodal Quality Assessment Framework for \textcolor{black}{Image Captioning} via Cross-Modal Signal Triangulation}

\begin{document}

\author{Ashim~Dahal$^{*1}$, Ankit~Ghimire$^{*1}$, Saydul~Akbar~Murad$^{2}$, Nick~Rahimi$^{1}$%
\thanks{$^{1}$Ashim Dahal, Ankit Ghimire, and Nick Rahimi are with the University of Southern Mississippi, Hattiesburg, MS, USA
(e-mail: \{ashim.dahal, ankit.ghimire, nick.rahimi\}@usm.edu).}%
\thanks{$^{2}$Saydul Akbar Murad is with the School of Business \& Technology, Emporia State University, Emporia, KS, USA
(e-mail: smurad@emporia.edu).}%
\thanks{$^{*}$ indicates equal contribution as co-first authors.}%
}

\maketitle

\begin{abstract}
Evaluating image captions requires cohesive assessment of both visual semantics and language pragmatics, which is often not entirely captured by most metrics. As such metrics increasingly guide model development, benchmarking, and system optimization in multimodal AI, inaccuracies in evaluation can misrepresent true progress. We introduce Redemption Score(RS), a novel evaluation framework for multi-modal generation by triangulating three complementary signals: (1) Mutual Information Divergence (MID) for global image-text distributional alignment, (2) DINO-based perceptual similarity of cycle-generated images for visual grounding, and (3) LLM Text Embeddings for contextual text similarity against human references. A calibrated fusion of these signals allows RS to offer a more holistic assessment. On the Flickr8k benchmark, RS achieves a Kendall-$\tau$ of 58.42, outperforming most prior methods and demonstrating superior correlation with human judgments without requiring task-specific training. Our framework provides a more robust and nuanced evaluation by thoroughly examining both the visual accuracy and text quality together, with consistent performance across Conceptual Captions and MS COCO.

\end{abstract}

\section{Introduction}

It is most commonly acknowledged that a picture is worth a thousand words, but the pragmatics of image captioning focus on finding the most plausible set of 10-20 words that best justifies describing its complexities. This precipitates a unique problem on linguistics deeply rooted to subjectivity. It dictates that two highly plausible captions grounded to the same image may lead to drastic degradation of traditional scoring metrics like BLEU \citep{papineni2002bleu}, BERTScore\citep{zhang2019bertscore}, METEOR \citep{banerjee2005meteor} or ROUGE \citep{lin2004rouge}. This motivates a principled reevaluation of how image captions are assessed, accounting for both visual semantics and linguistic interpretability.

Moreover, image captioning plays a crucial role in the development of multimodal systems, as it is widely used for benchmarking, model comparison, and guiding research progress in vision-language modeling. As a result, limitations in caption evaluation do not remain confined to the particular task, but directly influence how multimodal representations are learned, optimized, and compared.

Recent advancements on vision-grounded NLP, shared embedding spaces, and strong multimodal systems have dawned new possibilities for vision-grounded language, alongside a proliferation of evaluation metrics to rank image captions. These metrics can be grouped into three broad families:
\begin{itemize}
\item \textbf{Surface-overlap scores} (BLEU-4 \citep{papineni2002bleu}, METEOR \citep{banerjee2005meteor}, CIDEr \citep{vedantam2015cider} and SPICE \citep{anderson2016spice}): emphasize $n$-gram or scene-graph agreement yet can penalise legitimate paraphrases.
\item \textbf{Embedding-based or cross-modal measures} (BERTScore \citep{zhang2019bertscore}, CLIPScore \citep{hessel2021clipscore} and  Gaussian-assumed MID \citep{mid}): exploit pretrained representations but inherit modality biases and often struggle with fine-grained grounding.
\item \textbf{Cycle-consistent approaches} (CAMScore) \citep{camscore}: regenerate an image from the caption and compare it perceptually; sensitive to text-to-image artifacts and rely on a single synthetic perspective.
\end{itemize}

While each family illuminates part of the quality spectrum, none yields a fully reliable view of factual detail, linguistic nuance, and multi-reference visual grounding.

We introduce Redemption Score(RS), which preserves valuable visual information by combining three complementary evaluation signals: distributional alignment (how well image-text pairs fit learned representations), perceptual grounding (visual consistency through image regeneration), and linguistic fidelity (contextual text similarity). \textcolor{black}{RS uses general-purpose pretrained components, the current formulation
is not fully task-agnostic. In particular, the cycle-generation component assumes
that the model output is a descriptive caption that can be converted into a
proxy image through text-to-image generation. This assumption naturally fits
image captioning but does not directly transfer to tasks such as visual question
answering, image-text retrieval, or multimodal dialogue without task-specific
adaptation.} In summary, the key contributions of this paper are listed below:
\begin{itemize}
    \item A \textcolor{black}{model-training-free, reference-based} evaluation framework that addresses complementary failure modes in existing metrics, achieving high correlation with human judgments.
    \item An efficient calibration procedure that optimizes fusion weights through constrained grid search to maximize human alignment while ensuring meaningful contribution from all modalities.
    \item Demonstration of cross-dataset generalization from Flickr8k to Conceptual Captions and MS-COCO without parameter retuning, indicating robust transferability.
\end{itemize}

The formal definition of each component and the fusion rule are detailed in Section~\ref{sec:method}.

\section{Related Work}

\begin{table*}[!ht]
\footnotesize
\centering

\caption{Image–caption evaluation metrics grouped by family:
         \legendbox{surfB}{surface-overlap},
         \legendbox{embedB}{embedding/cross-modal},
         \legendbox{cycleB}{cycle/LLM},
         \legendbox{oursB}{our method}.}

\renewcommand{\arraystretch}{1.3}
\begin{tabularx}{\textwidth}{@{}
        L{1.9cm}  
        L{1.58cm} 
        L{1.18cm} 
        Y         
        Y         
        Y         
        @{}}
\toprule
\textbf{Metric} & \textbf{Type} & \textbf{Ref.} &
\textbf{Key Mechanism} & \textbf{Reported Strengths} &
\textbf{Limitations} \\ 
\midrule
\arrayrulecolor{black}

\rowcolor{surfA}
BLEU-4 \citep{papineni2002bleu} & N-gram                 & Text &
N-gram precision with brevity penalty &
Simplicity; fast computation &
Poor human correlation; ignores semantics; systematic bias \\

\rowcolor{surfA}
METEOR \citep{banerjee2005meteor} & N-gram + synonyms      & Text &
Unigram recall with stemming/synonym matching &
Better semantic grounding than BLEU &
Systematic bias; moderate agreement on long captions \\

\rowcolor{surfA}
CIDEr \citep{vedantam2015cider}  & N-gram + TF–IDF        & Text &
TF–IDF-weighted $n$-gram consensus &
Rewards reference consensus &
Negative model-level correlation; reference style sensitivity \\

\rowcolor{surfA}
SPICE \citep{anderson2016spice}  & Semantic Proposition   & Text &
Scene graph overlap (objects, attributes, relations) &
Evaluates semantic content &
Poor correlation for detailed captions; complex parsing \\
\addlinespace[2pt]

\rowcolor{embedA}
BERTScore  \citep{zhang2019bertscore}  & Textual embedding      & Text &
Cosine similarity of contextual BERT embeddings &
Captures paraphrase; good semantic alignment &
No visual grounding; misses factual errors \\

\rowcolor{embedA}
CLIPScore  \citep{hessel2021clipscore}  & Cross-modal            & None &
CLIP image–text embedding similarity &
Reference-free; image–text compatibility &
Modality gap; CLIP biases; fails on details \\

\rowcolor{embedA}
ViLBERTScore \citep{lee-etal-2020-vilbertscore} & Cross-modal            & Img + Text &
Image-conditioned ViLBERT embeddings &
Includes visual context &
Requires references; ViLBERT dependency \\

\rowcolor{embedA}
UMIC   \citep{lee2021umic}      & Cross-modal            & None &
UNITER with contrastive learning &
Reference-free; synthetic negatives &
Weak on counting/negation; detector needs \\

\rowcolor{embedA}
MID    \citep{mid}      & Cross-modal            & None &
CLIP feature mutual information &
Unified multimodal measure &
CLIP bias; Gaussian assumptions; costly \\

\rowcolor{embedA}
Polos   \citep{wada2024polos}     & Cross-modal (learned)  & Text (train) &
Multimodal feedback learning &
Learns hallucination cues &
Needs large feedback data; moderate performance \\

\rowcolor{embedA}
RefHICE-S  \citep{refhice-s} & Cross-modal (fine-grained) & Text (train) &
Hierarchical caption parsing + image alignment &
Captures fine-grained semantic alignment &
Complex parsing; new metric, higher computational cost \\

\rowcolor{embedA}
RefPAC-S++  \citep{refpac-S++} & Cross-modal            & Text (train) &
Reference-based PAC-S++ with multimodal feedback learning &
Combines reference consensus with image--text compatibility &
Needs large feedback data; degrades on non-COCO styles \\

\addlinespace[2pt]

\rowcolor{cycleA}
FLEUR/RefFLEUR  \citep{reffleur} & Cycle/LLM            & Text (train) + Image &
LMM-based scoring with reference captions and images; generates explanations &
High correlation with human judgment; explainable scores &
Requires large multimodal LLM; slower due to explanation generation \\

\rowcolor{cycleA}
VLM-as-Judge \citep{chen2024mllm} & LLM-based              & Varies &
LLM/VLM direct assessment &
High human correlation &
Judge bias; computational cost \\

\rowcolor{cycleA}
Image2Text2Image  \citep{image2text2image} & Cycle/LLM (round-trip) & None (reference-free) &
Text-to-image diffusion model round-trip: caption → diffusion-generated image compared to original via feature similarity &
Label-free evaluation; captures grammaticality, coverage, accuracy, hallucination detection; strong correlation with human judgment &
Depends on quality of text-to-image model, possibly diffusion biases; computationally expensive due to generation step \\

\rowcolor{cycleA}
CAMScore  \citep{camscore}   & Cycle-consistent       & None &
T2I generation + image comparison &
Avoids modality gap; reference-free &
T2I model bias; complex evaluation \\

\addlinespace[2pt]

\rowcolor{oursA}
\textbf{Our Method} & Hybrid fusion      & Img + Text &
MID + DINO + llm fusion &
High correlation; error detection &
Weight tuning; extra computations \\
\bottomrule
\end{tabularx}

\label{tab:metric_overview}
\end{table*}

To provide a structured overview of the field, Table~\ref{tab:metric_overview} summarizes representative image captioning evaluation metrics, grouped by methodological family, along with their core mechanisms and known limitations. This taxonomy highlights both the evolution of evaluation paradigms from surface level text overlap to multimodal, and cycle-consistent approaches along with the persistent gaps that motivate the proposed method.

\subsection{Historical Metrics}
These metrics were based on surface overlap and are best suited and designed to evaluate machine translation and text summarization and hold no information regarding the "image" component of image captioning. Metrics like BLEU, METEOR, CIDEr and SPICE \citep{papineni2002bleu, banerjee2005meteor, vedantam2015cider, anderson2016spice} fall under this category. The limitations for these metrics include but not limited to penalize re-phrasings and paraphrasing, and lack of visual grounding. 
\subsection{Embedding and Cross-Modal Metrics}
These are the second generation metrics based on multimodal or unimodal embeddings shift and similarities. Unimodal text-only variants include BERTScore, whereas multi-modal variants like CLIPScore, ViLBERTScore, UMIC, MID, and Polos \citep{zhang2019bertscore, hessel2021clipscore, lee-etal-2020-vilbertscore, lee2021umic, wada2024polos} metric share a common latent space for texts and images. Although these metrics improve semantic awareness, they still fail to overcome biases inherited from the encoder, fine-grained object attributes and inherit a persistent modality gap in which spatial relation and numeracy are encoded differently even in the shared vision and language channels.
\subsection{Cycle Consistent and Judge Models}
These are the most recent works in vision grounded NLP metrics which attempt to circumvent the modality gap in its entirety. The idea of using label free metric by employing the diffusion models to complete the image to text to image cycle was first introduced by Huang et al \citep{image2text2image}. Metrics like CAMScore \citep{camscore} regenerates an image from candidate caption via a text-to-image (T2I) model and compares the synthetic image to the original image using perceptual and object-level criteria. Likewise, VLM/LLM-as-a-Judge \cite{chen2024mllm, reffleur} frameworks employ the LLM foundation to assess caption quality directly through scalar scoring or pairwise preference. Furthermore, chain of thought reasoning was used by Tong et. al in \cite{tong2025g} to better evaluate the image captions for large multimodal models. Although these metrics achieve the highest known correlation with human judgements on long and detailed captions, cycle consistency inherits the biases and failures of T2I models, while the LLM-as-a-Judge models introduce heavy computational cost and calibration drift.

\subsection{Research Gap and Our Work}
Across the three diverse group of metrics, two key gaps are persistent: (i) \textbf{inclusive coverage}: no single metric captures global image to text alignment, local visual grounding and linguistic fidelity jointly and (ii) \textbf{robustness}: metrics often inherit systematic biases from their underlying models and reference corpora leading to misleading ranking in exception cases. To address these gaps, we propose Redemption Score - a \textcolor{black}{model-training-free, reference-based} framework that mitigates individual metric biases through complementary signal integration. Our approach combines MID for distributional alignment, DINO similarity \citep{Caron2021EmergingPI} for visual consistency through cycle-generated images, and general text embeddings for semantic understanding.

\begin{figure*}[!ht]
    \centering
    \includegraphics[width=\linewidth]{figures/compressed.pdf}
    \caption{Overview of calculation of Redemption Score.}
    \label{fig:methodology}
\end{figure*}

\section{Redemption Score}\label{sec:method}
An overview of calculation of Redemption Score is highlighted on \cref{fig:methodology} which is in-detail described in the following sections.

These three components address different failure modes in caption evaluation: MID captures whether image-caption pairs follow expected distributional patterns (detecting statistical outliers), DINO similarity identifies visual inconsistencies through cycle generation, and GTE embeddings catch semantic mismatches that surface-level metrics miss. No single component can reliably detect all caption quality issues.

\subsection{Mutual Information Divergence (MID)}

\label{sec:mid_concise}

We adapt the Mutual Information Divergence (MID) from \citet{mid} to capture both image semantics and language pragmatics. Let $g(I)$ and $h(\hat{c})$ be the $\ell_{2}$-normalized CLIP ViT-L/14 \cite{clip} embeddings of an image $I$ and a candidate caption $\hat{c}$. Assuming these embeddings (random variables $X$ for images, $Y$ for captions) follow multivariate Gaussian distributions with means $\mu_x, \mu_y$ and covariances $\Sigma_x, \Sigma_y$, the continuous mutual information is:
\begin{equation}
I(X;Y) = \tfrac{1}{2}\log \frac{\det(\Sigma_x)\det(\Sigma_y)}{\det(\Sigma_{xy})}
\label{eq:cmi_concise}
\end{equation}
where $\Sigma_{xy}$ is the joint covariance.

\citet{mid} define a point-wise mutual information (PMI) for individual pairs $(x,y)$, which adjusts $I(X;Y)$ using Mahalanobis distances to reflect how well the pair aligns with the learned distributions. We redefine the PMI normalization on \cref{eq:norm}. The final MID score is the expectation of these PMI values:
\begin{equation}
\operatorname{MID}(I,\hat{c}) = \mathbb{E}_{(x,y)\sim p}[\operatorname{PMI}(x,y)]
\label{eq:mid_concise}
\end{equation}
The necessary distributional statistics $(\mu_x,\Sigma_x,\mu_y,\Sigma_y, \Sigma_{xy})$ are estimated once from the entire dataset (details in \S\ref{sec:exp_setup}).

\subsection{DINO Perceptual Similarity}
\label{sec:dino}
The perceptual component leverages the self-supervised
\textsc{DINO} vision transformer \citep{Caron2021EmergingPI}. This component captures visual consistency through cycle generation and comparison.
For every example we consider the original image \(I\), its human reference caption \(c^{\text{ref}}\), and the model-generated candidate caption \(\hat{c}\). Using the public Stable Diffusion 3 checkpoint \citep{stablediffusion}, we synthesize two proxy images $\tilde{I}_{\text{ref}}$ and $\tilde{I}_{\text{cand}}$.
All images are resized to \(224\times224\) and
ImageNet-normalised before encoding.
\paragraph{Feature extraction.}
A frozen \texttt{ViT-B/8\,DINO} encoder
\(E(\cdot)\in\mathbb{R}^{768}\) produces a \texttt{[CLS]} embedding
for every image.  Embeddings are $\ell_{2}$-normalised:
\(\hat{e}(X)=E(X)/\lVert E(X)\rVert_{2}\).

\paragraph{Cosine similarities.}
We compute two edge scores
\begin{align}
s_{1} &= \bigl\langle \hat{e}(I),\,
                     \hat{e}(\tilde I_{\mathrm{cand}})\bigr\rangle,\\
s_{2} &= \bigl\langle \hat{e}(\tilde I_{\mathrm{cand}}),\,
                     \hat{e}(\tilde I_{\mathrm{ref}})\bigr\rangle
\end{align}

\paragraph{Aggregate perceptual score.}
The final RS receives the mean of these two scores.
\begin{equation}
\mathrm{DINO_{sim}}
\;=\;
\tfrac12\bigl(s_{1}+s_{2}\bigr)
\end{equation}

\subsection{LLM Embeddings}
\label{sec:llm}

We adopt General Text Embeddings \citep{li2023towards} with the
thenlper/gte-large encoder to account for language pragmatics.
Let \(\mathbf{e}(x)\in\mathbb{R}^d\) denote the (L2-normalized) GTE embedding
of text \(x\) produced by the thenlper/gte-large encoder. For a candidate caption
\(\hat c\) and its reference caption \(c^{\mathrm{ref}}\), we compute the GTEScore as the cosine similarity between their embeddings:
\begin{equation}
\mathrm{GTEScore}
\;=\;
\cos\!\bigl(\mathbf{e}(\hat c),\,\mathbf{e}(c^{\mathrm{ref}})\bigr)
\;=\;
\frac{\mathbf{e}(\hat c)^\top \mathbf{e}(c^{\mathrm{ref}})}
{\|\mathbf{e}(\hat c)\|_2 \,\|\mathbf{e}(c^{\mathrm{ref}})\|_2}
\end{equation}

\subsection{Redemption Score Aggregation}
\label{sec:redemption_score_aggregration}
\paragraph{Normalization.}
To ensure all three signals are directly comparable and compatible with the multiplicative component, we normalize all scores to the unit interval $[0,1]$ using the transformation:
\begin{equation}
    \label{eq:norm}
    X_{normalized} = \frac{\frac{X}{1+|X|} + 1}{2}
\end{equation}
where $X$ represents the MID component from (Equation~\ref{eq:mid_concise}), the raw DINO similarity score, or the GTE cosine similarity score. This parameter-free transformation avoids the need for dataset-dependent statistics (as required by min-max or z-score normalization), making the metric practical for evaluating individual captions or small caption sets without requiring pre-computed normalization parameters. The transformation ensures all signals are positive and bounded in [0,1], with larger input values mapping to larger output values, enabling meaningful aggregation across the three modalities.

\noindent
\paragraph{Aggregation.}

The final \emph{Redemption Score (RS)} uses a hybrid aggregation that interpolates between linear combination and weighted geometric mean of the three normalized signals, with learnable weights $\alpha$, $\beta$, $\gamma$ and interpolation parameter $\lambda$. This hybrid approach combines the robustness of additive aggregation with the strict quality control of multiplicative aggregation, providing the benefits of both strategies.

\begin{align}
\label{eq:redeem}
    RS[i] &= \lambda \cdot L[i] + (1-\lambda) \cdot M[i]
\end{align}
where:
\begin{itemize}
    \item $L[i] = \alpha Z_{mid}[i] + \beta Z_{dino}[i] + \gamma Z_{gtescore}[i]$ (linear component)
    \item $M[i] = Z_{mid}[i]^{\alpha} Z_{dino}[i]^{\beta} Z_{gtescore}[i]^{\gamma}$ (multiplicative component)
\end{itemize}
with constraints $\alpha+\beta+\gamma=1$ and $\alpha,\beta,\gamma > 0$.

\paragraph{Parameter optimization.}
We employ a constrained optimization approach to determine optimal parameters for our proposed metric. Our methodology addresses two key objectives: (1) maximizing correlation with human judgments, and (2) ensuring meaningful contribution from all metric components.

We implement minimum weight constraints ($\alpha, \beta, \gamma \geq 0.15$) to prevent degenerate solutions where any single modality dominates the final score. This threshold ensures that each modality--- distributional alignment(MID), visual similarity (DINO), and contextual understanding (GTE embeddings)---contributes meaningfully to the evaluation while allowing data-driven optimization to determine their relative importance.

The constrained grid search explores the parameter space with step size $\Delta = 0.05$ for weights and $\Delta = 0.1$ for the interpolation parameter $\lambda$.

\paragraph{Parameter selection methodology.}
We select parameters using the following criteria:

\textbf{Optimization objective:} We identify the parameter combination that maximizes Kendall's $\tau$ correlation with human judgments while satisfying the minimum weight constraint ($\alpha, \beta, \gamma \geq 0.15$).

\textbf{Statistical significance:} All selected parameters must achieve statistical significance ($p < 0.05$) in their correlation with human ratings.

\textbf{Robustness validation:} We perform sensitivity analysis around the optimal parameters to ensure stable performance under small perturbations.

The resulting parameter configuration reflects the empirical evidence for optimal modality weighting as determined by human evaluation data, constrained by methodological requirements for multi-modal contribution.

\begin{figure*}[t!]
    \centering
    \subfloat[{\footnotesize \textcolor{black}{Optimal values of $\alpha, \beta, \gamma, \lambda$ maximizing Kendall-$\tau$.}}\label{fig:abg_optimal_sub}]{%
        \includegraphics[width=0.32\textwidth]{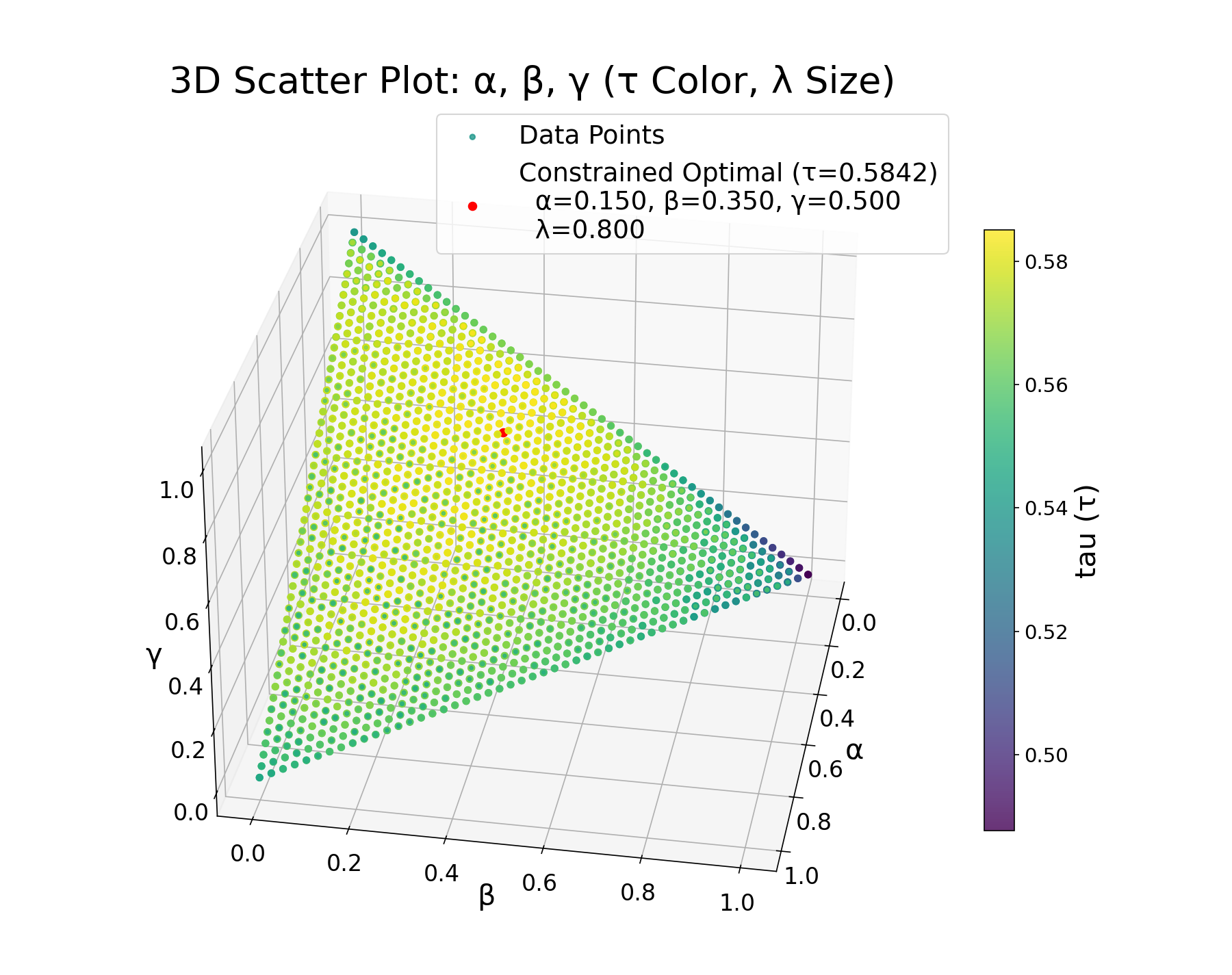}%
    }
    \hfill
    \subfloat[{\footnotesize RS vs. Kendall-$\tau$ relationship with fitted models.}\label{fig:kendall-tau-rs-relationship_sub}]{%
        \includegraphics[width=0.37\textwidth]{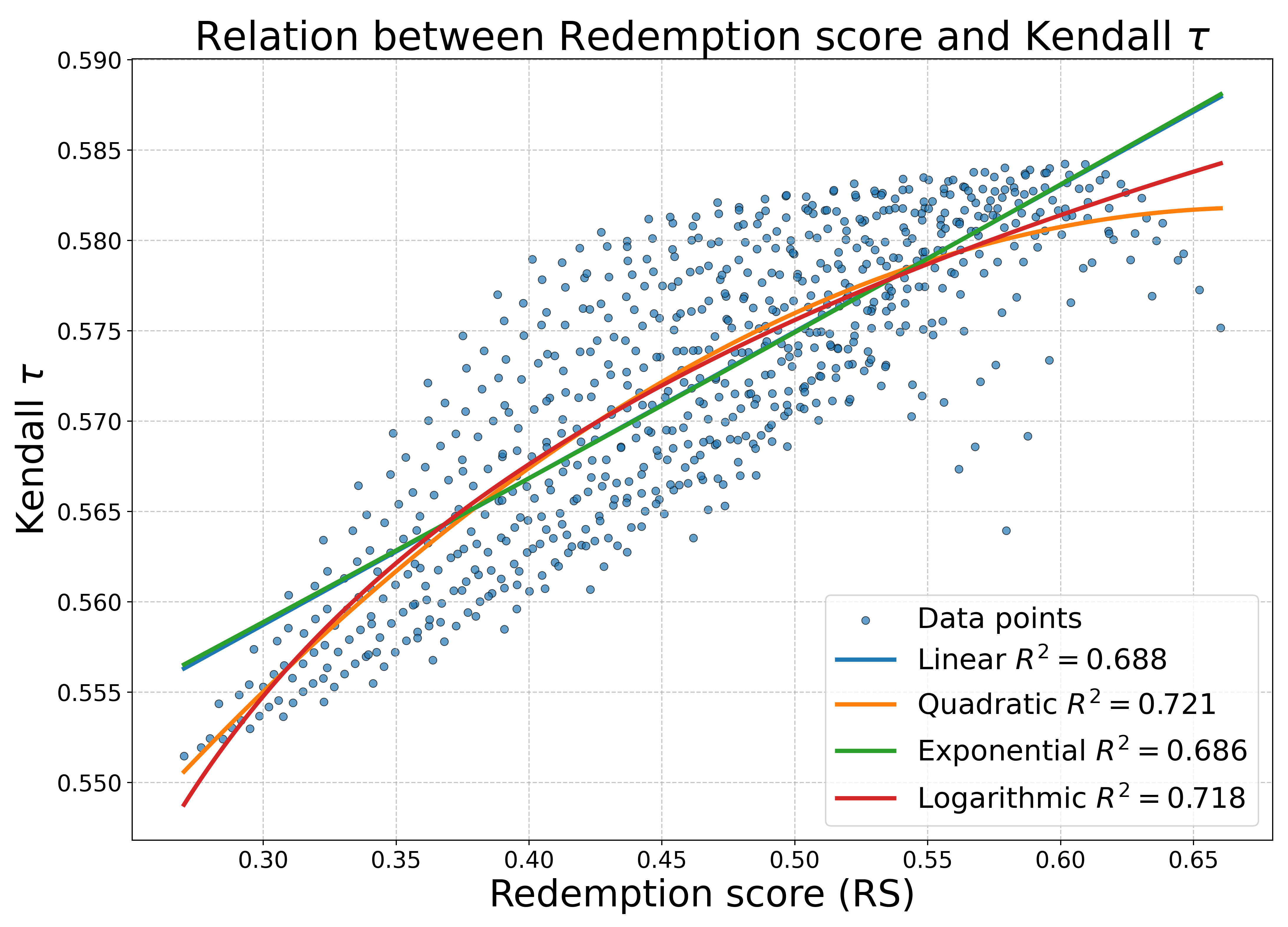}%
    }
    \hfill
    \subfloat[{\footnotesize Comparative radar plot of various metrics' scores and Kendall-$\tau$.}\label{fig:metrics_radar_plot_sub}]{%
        \includegraphics[width=0.27\textwidth]{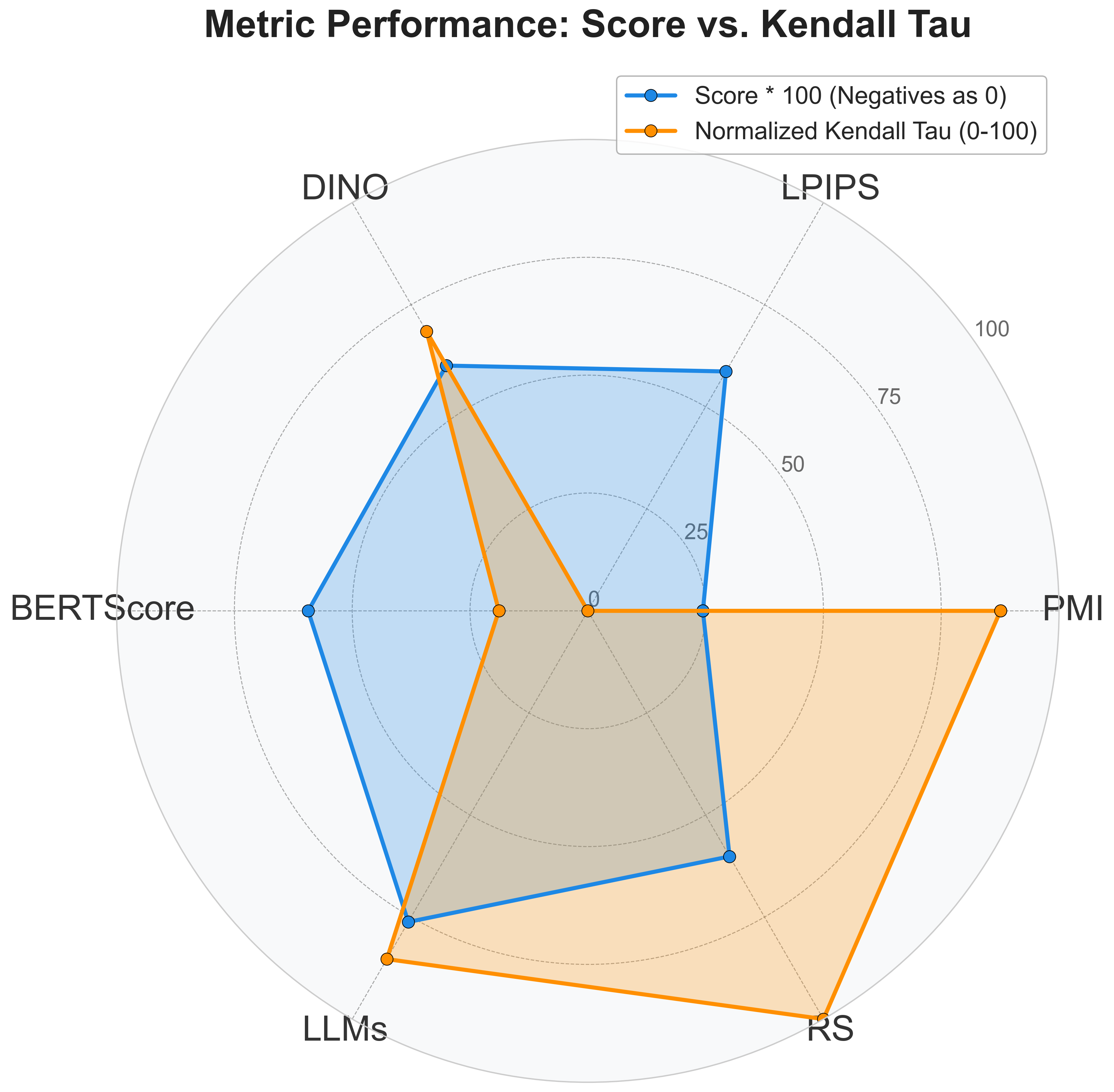}%
    }
    
    \caption{Comprehensive analysis of Redemption Score (RS) parameters, its relationship with Kendall-$\tau$, and comparison with other metrics. Comparison with more metrics is shown on \cref{tab:f8k_comparison}}
    \label{fig:combined_analysis_three_plots}
\end{figure*}
\section{Experiments}
\subsection{Dataset}\label{sec:exp_setup}

We optimized our Redemption Score on the Flickr8k dataset \citep{flickr}, using its full validation set (\(5{,}822\) images) and human preference data to maximize Kendall-$\tau$ correlation. We exclude 158 image-caption pairs where candidate captions were exactly the reference captions. This process yielded the score's weights \((\alpha,\beta,\gamma,\lambda)\) (see \cref{eq:redeem}). To evaluate generalizability, we applied these same weights to MS-COCO \citep{coco} and Conceptual Captions \citep{sharma2018conceptual} without further tuning. While MS-COCO has human quality ratings that could enable dataset-specific optimization, we prioritized testing the metric's ability to transfer across datasets with a single set of weights. Conceptual Captions lacks human preference data necessary for Kendall-$\tau$ optimization.

\vspace{1.5em}

\noindent\textbf{Workload Summary.} The DINO Similarity component required generating approximately \(36{,}000\) proxy images (1 reference + 5 candidates generated by models in \cref{sec:captioning_models} for \(3,000\) samples across the 2 datasets). Adding the base image comparisons, DINO similarity processed around \(42{,}000\) images. In contrast, MID operated directly on image-caption pairs, resulting in roughly \(60{,}000\) total comparisons: (\(40{,}000\) from the full Flickr8k split and \(20{,}000\) from the Conceptual Captions and COCO validation subsets). GTEScore operated on the same captions as DINO Similarity resulting to \(36{,}000\) total text-text comparisons.

\begin{table*}[t]
\footnotesize                
\setlength{\tabcolsep}{2pt}
\renewcommand{\arraystretch}{1.2}

\centering
\caption{Metric values and Kendall–$\tau_c$ correlation on raw scores without normalization on Flickr8k dataset.}
\label{tab:metric_single}

\begin{tabularx}{\linewidth}{l
  *{2}{>{\centering\arraybackslash}X}
  *{2}{>{\centering\arraybackslash}X}
  *{2}{>{\centering\arraybackslash}X}
  *{2}{>{\centering\arraybackslash}X}
  *{2}{>{\centering\arraybackslash}X}
  *{2}{>{\centering\arraybackslash}X}
  *{2}{>{\centering\arraybackslash\columncolor{oursA}}X}}
\toprule
 & \multicolumn{2}{c}{BLEU-4}
 & \multicolumn{2}{c}{METEOR}
 & \multicolumn{2}{c}{MID}
 & \multicolumn{2}{c}{DINO}
 & \multicolumn{2}{c}{BERTScore}
 & \multicolumn{2}{c}{GTEScore}
& \multicolumn{2}{c}{\cellcolor{oursA}\textbf{RS}} \\
\cmidrule(lr){2-3}\cmidrule(lr){4-5}\cmidrule(lr){6-7}
\cmidrule(lr){8-9}\cmidrule(lr){10-11}\cmidrule(lr){12-13}
Dataset & Val & $\tau$ & Val & $\tau$ & Val & $\tau$ & Val & $\tau$
        & Val & $\tau$ & Val & $\tau$ & Val & $\tau$ \\
\midrule
Flickr8k &
0.0465 & 33.5 &
0.2441 & 35.85 &
$-17.55$ & \textbf{54.6} &
0.268 & 48.76 &
0.59 & 38.05 &
0.76 & 53.93 &
0.48 & \textbf{58.4} \\
\bottomrule
\end{tabularx}
\end{table*}


\begin{table*}[t]
\footnotesize
\setlength{\tabcolsep}{2pt}
\renewcommand{\arraystretch}{1.2}
\centering
\caption{Metric scores per captioning model across Conceptual Captions and COCO.
 RS uses fixed $(\alpha,\beta,\gamma,\lambda)=(0.15,0.35,0.5,0.8)$ (tuned on Flickr8k).}
\label{tab:model_metrics_combined}
\renewcommand{\arraystretch}{1.3}
\begin{tabular}{
 >{\raggedright\arraybackslash}p{2.0cm} 
 >{\centering\arraybackslash}p{1.3cm} 
 >{\centering\arraybackslash}p{1.2cm} 
 >{\centering\arraybackslash}p{1.6cm} 
 >{\centering\arraybackslash}p{1.2cm} 
 >{\centering\arraybackslash}p{1.2cm} 
 >{\centering\arraybackslash}p{1.2cm} 
 >{\centering\arraybackslash}p{.8cm} 
 >{\centering\arraybackslash}p{1.7cm} 
 >{\centering\arraybackslash}p{1.7cm} 
 >{\centering\arraybackslash\columncolor{oursA}}p{1.8cm} 
}

\toprule
\textbf{Model} & \textbf{METEOR} & \textbf{CIDEr} & \textbf{ROUGE-L} &
\textbf{SPICE} & \textbf{CLIP-S} & \textbf{DINO} & \textbf{MID} &
\textbf{BERTScore} & \textbf{GTEScore} & \textbf{RS} \\ \midrule

\rowcolor{surfB}\multicolumn{11}{l}{Conceptual Captions Dataset}\\[2pt]

Qwen2-VL-7B & 0.10 & 0.05 & 0.14 & 0.08 & 0.64 & 0.52 & 30.0 & 0.52 & 0.85 & 0.80\\
GIT (base) & 0.07 & 0.36 & 0.20 & 0.12 & 0.60 & 0.38 & 17.0 & 0.47 & 0.83 & 0.77\\
ViT-GPT2 & 0.07 & 0.25 & 0.17 & 0.09 & 0.59 & 0.39 & 12.7 & 0.53 & 0.81 & 0.74\\
BLIP & 0.12 & 0.98 & 0.28 & 0.20 & 0.62 & 0.48 & 30.5 & 0.60 & 0.86 & 0.81\\
BLIP-2 & 0.13 & 0.97 & 0.29 & 0.20 & 0.63 & 0.49 & 32.3 & 0.60 & 0.87 & 0.81\\

\rowcolor{surfB}\multicolumn{11}{l}{COCO Dataset}\\[2pt]

Qwen2-VL-7B & 0.16 & 0.01 & 0.21 & 0.11 & 0.64 & 0.60 & 30.2 & 0.60 & 0.88 & 0.82\\
GIT (base) & 0.07 & 0.29 & 0.19 & 0.13 & 0.60 & 0.44 & 14.3 & 0.47 & 0.84 & 0.75\\
ViT-GPT2 & 0.15 & 0.82 & 0.34 & 0.21 & 0.61 & 0.55 & 26.8 & 0.65 & 0.87 & 0.81\\
BLIP & 0.12 & 0.69 & 0.28 & 0.21 & 0.62 & 0.56 & 29.6 & 0.61 & 0.87 & 0.81\\
BLIP-2 & 0.12 & 0.59 & 0.26 & 0.18 & 0.63 & 0.56 & 29.9 & 0.57 & 0.87 & 0.81\\
\bottomrule
\end{tabular}
\end{table*}

\subsection{Captioning Systems}\label{sec:captioning_models}
The Conceptual Captions dataset was used to QLoRA \citep{Dettmers2023QLoRAEF} fine-tune 5 popular multimodal systems: (i) BLIP \citep{Li2022BLIPBL}, (ii) BLIP2 2.7B \citep{Li2023BLIP2BL}, (iii) MS-GIT \citep{Wang2022GITAG}, (iv) ViT-GPT-2 \citep{vit, gpt2} and (v) Qwen 2.5-VL 7B \citep{wang2024qwen2}. 
We trained captioning models on the Conceptual Captions dataset, which is sourced from web alt-text and thus provides captions with greater stylistic diversity and less rigidly descriptive tendencies, offering broader generalization potential but also posing challenges for methods assuming literal image–text alignment \citep{sharma2018conceptual,ng2020understanding}. 
All models were fine-tuned for one epoch with an effective learning
rate of 5e-5, QLoRA rank of 8 and quantization of 4-bits on their respective predefined-loss functions.
We observed that larger LLM-style models such as Qwen tended to generate meta-statements (e.g., “I can think of a few different captions”) and enumerate multiple caption options, which is undesirable for standard captioning tasks. To mitigate this behavior, we fine-tuned the larger models, and to ensure fairness across baselines, we applied the same fine-tuning procedure to all captioning systems.


\begin{table}[!t] 
\footnotesize 
\centering
\caption{Flickr8K-Expert scores and qualitative properties.
$\Delta$ indicates the difference from our Redemption score
($\tau=58.4$). Positive values (green) indicate our score is higher. RT=Requires Training, IG=Image Grounded}
\label{tab:f8k_comparison}
\renewcommand{\arraystretch}{1.3}
\begin{tabular}{
 p{2.9cm} 
 C{1 cm} 
 C{.8 cm} 
 C{.8 cm} 
 C{1cm} 
}
\toprule
\bfseries Metric & 
\bfseries \shortstack{Flickr8K\\(Expert)} & 
\bfseries \shortstack{RT} & 
\bfseries \shortstack{IG} & 
$\boldsymbol{\Delta}$\\
\midrule
ROUGE \citep{lin2004rouge} & 32.3 & \xmark & \xmark & \textcolor{ForestGreen}{(+26.1)} \\
CIDEr \citep{vedantam2015cider} & 43.9 & \xmark & \xmark & \textcolor{ForestGreen}{(+14.5)} \\
SPICE \citep{anderson2016spice} & 44.9 & \xmark & \xmark & \textcolor{ForestGreen}{(+13.5)} \\
SPARCS \citep{feinglass-yang-2021-smurf} & 48.1 & \xmark & \xmark & \textcolor{ForestGreen}{(+10.3)} \\
MoverScore \citep{Zhao2019MoverScoreTG} & 46.7 & \xmark & \xmark & \textcolor{ForestGreen}{(+11.7)} \\
BARTScore \citep{bartscore_e4d2b6e6} & 37.8 & \xmark & \xmark & \textcolor{ForestGreen}{(+20.6)} \\
UMIC \citep{lee2021umic} & 46.8 & \cmark & \cmark & \textcolor{ForestGreen}{(+11.6)} \\
ViLBERTScore \citep{lee-etal-2020-vilbertscore} & 50.1 & \cmark & \cmark & \textcolor{ForestGreen}{(+8.3)} \\
CLIP-S \citep{hessel2021clipscore} & 51.2 & \xmark & \cmark & \textcolor{ForestGreen}{(+7.2)} \\
RefCLIP-S \citep{hessel2021clipscore} & 53.0 & \xmark & \cmark & \textcolor{ForestGreen}{(+5.4)} \\
FLUER \citep{reffleur} & 53.0 & \xmark & \cmark & \textcolor{ForestGreen}{(+5.4)} \\
PAC-S \citep{Sarto2023PositiveAugmentedCL} & 54.3 & \cmark & \cmark & \textcolor{ForestGreen}{(+4.1)} \\
RefPAC-S \citep{Sarto2023PositiveAugmentedCL} & 55.9 & \cmark & \cmark & \textcolor{ForestGreen}{(+2.5)} \\
Polos \citep{wada2024polos} & 56.4 & \xmark & \cmark & \textcolor{ForestGreen}{(+2.0)} \\
DENEB \citep{matsuda2024deneb} & 56.8 & \cmark & \cmark & \textcolor{ForestGreen}{(+1.6)} \\
RefHICE-S \citep{zeng2024hicescore} & 57.7 & \xmark & \cmark & \textcolor{ForestGreen}{(+0.7)} \\
G-VEval-ref-free \citep{tong2025g} & 59.7 & \xmark & \cmark & \textcolor{red}{(-1.3)} \\
\textcolor{black}{CAMScore \citep{camscore}} & 55.6 & \xmark & \cmark & \textcolor{ForestGreen}{(+2.8)} \\
\textcolor{black}{EXPERT \citep{kim2025expert}} & 56.7 & \cmark & \cmark & \textcolor{ForestGreen}{(+1.7)} \\
\textcolor{black}{DISCODE-IN \citep{inoue2026discode}} & 58.1 & \xmark & \cmark & \textcolor{ForestGreen}{(+0.3)} \\
\textcolor{black}{Pearl \citep{hirano2026pearl}} & 58.6 & \cmark & \cmark & \textcolor{red}{(-0.2)} \\
\midrule
\rowcolor{oursA} 
\bfseries RS (ours) & 
\bfseries 58.4 &
\xmark & \cmark & --- \\ 
\bottomrule
\end{tabular}
\end{table}

\subsection{Results}
\paragraph{Finding $\alpha, \beta, \gamma, \lambda$.} The optimal values of $\alpha, \beta, \gamma, \lambda$ were found to be 0.15, \textcolor{black}{0.35, 0.50}, and 0.80 respectively via grid search as described in \cref{sec:redemption_score_aggregration}. The grid search and optimal value is highlighted in \cref{fig:abg_optimal_sub}. \textcolor{black}{The fusion weights $(\alpha,\beta,\gamma,\lambda)
=(0.15,0.35,0.50,0.80)$ are selected exclusively on Flickr8k
and are subsequently frozen for all evaluations, including
MS-COCO, Pascal-50S, and Nebula; no dataset-specific retuning
is performed.
}

\paragraph{Results on Flickr8k.} \Cref{tab:metric_single}, \cref{tab:f8k_comparison} and \cref{fig:combined_analysis_three_plots} summarize the findings on the Flickr8k dataset. Since the Flickr8k dataset contains random mapping of captions to images, many images are paired with unrelated captions, thus achieving lower human ratings. \Cref{fig:metrics_radar_plot_sub}, \cref{fig:kendall-tau-rs-relationship_sub} and \cref{tab:metric_single} demonstrate that beyond a certain threshold, higher metric scores correspond to worse Kendall-$\tau$ correlations with human evaluators, indicating that extremely high automatic scores may reflect semantic misalignment rather than quality. Overall, we achieved a Kendall-$\tau$ score of 58.42 on the dataset, which surpasses traditional and most recent methods in \cref{tab:metric_single}.

\begin{figure*}[htbp]
\centering
\caption{Complementary failure modes captured by different components. Each row shows cases where one component fails to detect caption quality issues that human raters identify, while our integrated RS Score successfully captures the problem through its multi-modal approach. Human ratings: 1-4 scale (higher = better quality), Metric scores: 0-1 scale (higher = better alignment).}
\label{fig:metric_failures}

\noindent\textbf{MID Score Failures} \\
\textit{\scriptsize MID gives unexpectedly low scores to captions that humans rate as reasonable, suggesting limitations in its distributional modeling approach}
\vspace{0.1cm}

\begin{minipage}[t]{0.31\textwidth}
 \centering
 \begin{tabular}{@{}c@{\hspace{0.1cm}}c@{}}
 \fbox{\includegraphics[width=0.5\textwidth, height=2.5cm, keepaspectratio]{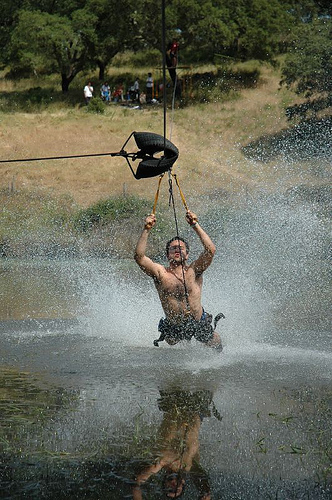}} &
 \raisebox{0.8cm}{%
 \tiny
 \begin{tabular}{|l|c|}
 \hline
 MID & \textcolor{black}{\textbf{0.081}} \\
 \hline
 RS & \textcolor{teal}{\textbf{0.649}} \\
 \hline
 \textbf{Human} & \textbf{2.67} \\
 \hline
 \end{tabular}%
 }
 \end{tabular}
 \vspace{0.1cm}\\
 \scriptsize
 \textbf{Caption:} Man falling off a blue surfboard in the ocean.
\end{minipage}%
\hfill%
\begin{minipage}[t]{0.31\textwidth}
 \centering
 \begin{tabular}{@{}c@{\hspace{0.1cm}}c@{}}
 \fbox{\includegraphics[width=0.5\textwidth, height=2.5cm, keepaspectratio]{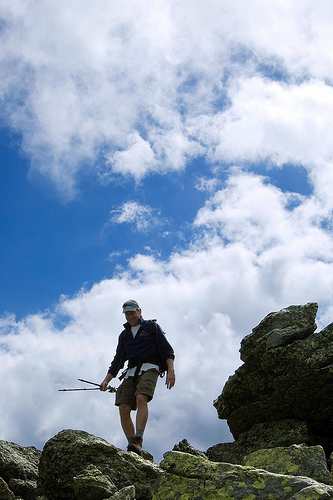}} &
 \raisebox{0.8cm}{%
 \tiny
 \begin{tabular}{|l|c|}
 \hline
 MID & \textcolor{black}{\textbf{0.114}} \\
 \hline
 RS & \textcolor{teal}{\textbf{0.634}} \\
 \hline
 \textbf{Human} & \textbf{3.00} \\
 \hline
 \end{tabular}%
 }
 \end{tabular}
 \vspace{0.1cm}\\
 \scriptsize
 \textbf{Caption:} A rock climber climbs in between two very large rocks.
\end{minipage}%
\hfill%
\begin{minipage}[t]{0.31\textwidth}
 \centering
 \begin{tabular}{@{}c@{\hspace{0.1cm}}c@{}}
 \fbox{\includegraphics[width=0.5\textwidth, height=2.5cm, keepaspectratio]{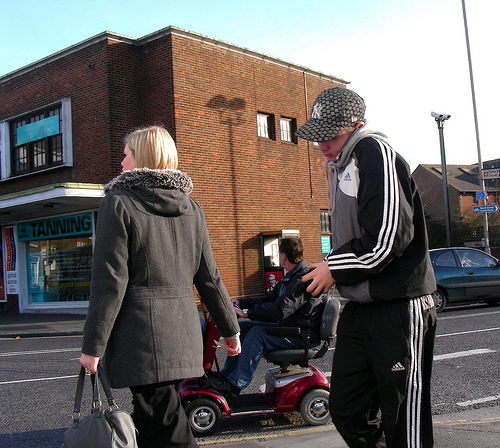}} &
 \raisebox{0.8cm}{%
 \tiny
 \begin{tabular}{|l|c|}
 \hline
 MID & \textcolor{black}{\textbf{0.084}} \\
 \hline
 RS & \textcolor{teal}{\textbf{0.593}} \\
 \hline
 \textbf{Human} & \textbf{2.67} \\
 \hline
 \end{tabular}%
 }
 \end{tabular}
 \vspace{0.1cm}\\
 \scriptsize
 \textbf{Caption:} A man standing in front of a brick building.
\end{minipage}

\vspace{0.3cm}

\noindent\textbf{DINO Score Failures} \\
\textit{\scriptsize DINO penalizes captions too harshly for minor visual inaccuracies or missing details that humans consider acceptabley}
\vspace{0.1cm}

\begin{minipage}[t]{0.31\textwidth}
 \centering
 \begin{tabular}{@{}c@{\hspace{0.1cm}}c@{}}
 \fbox{\includegraphics[width=0.5\textwidth, height=2.5cm, keepaspectratio]{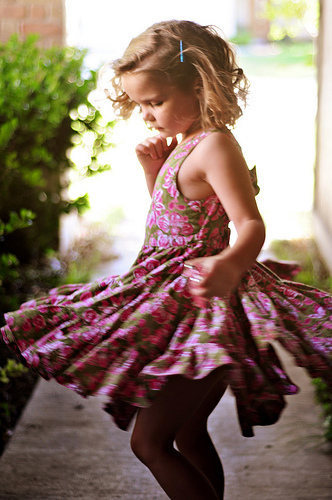}} &
 \raisebox{0.8cm}{%
 \tiny
 \begin{tabular}{|l|c|}
 \hline
 DINO & \textcolor{black}{\textbf{0.182}} \\
 \hline
 RS & \textcolor{teal}{\textbf{0.746}} \\
 \hline
 \textbf{Human} & \textbf{3.00} \\
 \hline
 \end{tabular}%
 }
 \end{tabular}
 \vspace{0.1cm}\\
 \scriptsize
 \textbf{Caption:} A young girl is wearing a purple shirt and pink headband.
\end{minipage}%
\hfill%
\begin{minipage}[t]{0.31\textwidth}
 \centering
 \begin{tabular}{@{}c@{\hspace{0.1cm}}c@{}}
 \fbox{\includegraphics[width=0.5\textwidth, height=2.5cm, keepaspectratio]{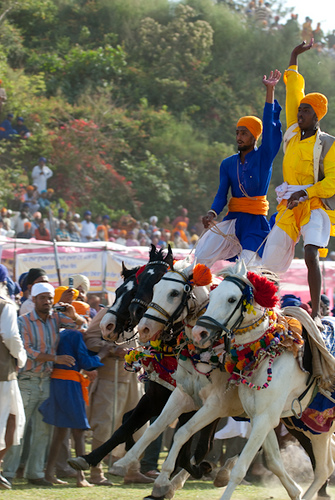}} &
 \raisebox{0.8cm}{%
 \tiny
 \begin{tabular}{|l|c|}
 \hline
 DINO & \textcolor{black}{\textbf{0.177}} \\
 \hline
 RS & \textcolor{teal}{\textbf{0.699}} \\
 \hline
 \textbf{Human} & \textbf{3.00} \\
 \hline
 \end{tabular}%
 }
 \end{tabular}
 \vspace{0.1cm}\\
 \scriptsize
 \textbf{Caption:} A crowd of people at an outdoor event.
\end{minipage}%
\hfill%
\begin{minipage}[t]{0.31\textwidth}
 \centering
 \begin{tabular}{@{}c@{\hspace{0.1cm}}c@{}}
 \fbox{\includegraphics[width=0.5\textwidth, height=2.5cm, keepaspectratio]{images/2077079696_03380d218b.jpg}} &
 \raisebox{0.8cm}{%
 \tiny
 \begin{tabular}{|l|c|}
 \hline
 DINO & \textcolor{black}{\textbf{0.114}} \\
 \hline
 RS & \textcolor{teal}{\textbf{0.697}} \\
 \hline
 \textbf{Human} & \textbf{2.67} \\
 \hline
 \end{tabular}%
 }
 \end{tabular}
 \vspace{0.1cm}\\
 \scriptsize
 \textbf{Caption:} Two men holding their fishing poles.
\end{minipage}

\vspace{0.3cm}

\noindent\textbf{GTEScore Failures} \\ 
\textit{\scriptsize General Text Embeddings overweight linguistic fluency, missing visual-semantic misalignments}
\vspace{0.1cm}

\begin{minipage}[t]{0.31\textwidth}
 \centering
 \begin{tabular}{@{}c@{\hspace{0.1cm}}c@{}}
 \fbox{\includegraphics[width=0.5\textwidth, height=2.5cm, keepaspectratio]{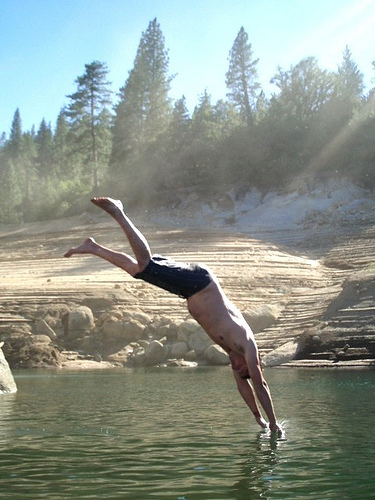}} &
 \raisebox{0.8cm}{%
 \tiny
 \begin{tabular}{|l|c|}
 \hline
 GTEScore & \textcolor{black}{\textbf{0.664}} \\
 \hline
 RS & \textcolor{teal}{\textbf{0.495}} \\
 \hline
 \textbf{Human} & \textbf{1.67} \\
 \hline
 \end{tabular}%
 }
 \end{tabular}
 \vspace{0.1cm}\\
 \scriptsize
 \textbf{Caption:} Two dogs are jumping up at each other.
\end{minipage}%
\hfill%
\begin{minipage}[t]{0.31\textwidth}
 \centering
 \begin{tabular}{@{}c@{\hspace{0.1cm}}c@{}}
 \fbox{\includegraphics[width=0.5\textwidth, height=2.5cm, keepaspectratio]{images/3692892751_f6574e2700.jpg}} &
 \raisebox{0.5cm}{%
 \tiny
 \begin{tabular}{|l|c|}
 \hline
 GTEScore & \textcolor{black}{\textbf{0.709}} \\
 \hline
 RS & \textcolor{teal}{\textbf{0.500}} \\
 \hline
 \textbf{Human} & \textbf{1.67} \\
 \hline
 \end{tabular}%
 }
 \end{tabular}
 
 \vspace{0.45cm}
 
 \scriptsize
 \textbf{Caption:} Three young adults look towards the camera in a school setting.
\end{minipage}%
\hfill%
\begin{minipage}[t]{0.31\textwidth}
 \centering
 \begin{tabular}{@{}c@{\hspace{0.1cm}}c@{}}
 \fbox{\includegraphics[width=0.5\textwidth, height=2.5cm, keepaspectratio]{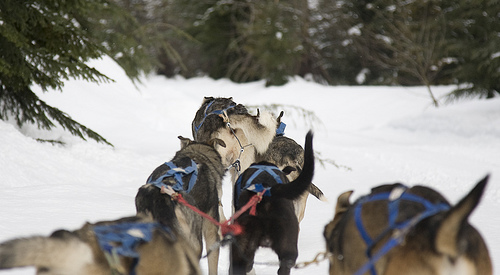}} &
 \raisebox{0.3cm}{%
 \tiny
 \begin{tabular}{|l|c|}
 \hline
 GTEScore & \textcolor{black}{\textbf{0.708}} \\
 \hline
 RS & \textcolor{teal}{\textbf{0.490}} \\
 \hline
 \textbf{Human} & \textbf{1.00} \\
 \hline
 \end{tabular}%
 }
 \end{tabular}
 
 \vspace{0.6cm}
 
 \scriptsize
 \textbf{Caption:} A little boy slides down a bright red corkscrew slide.
\end{minipage}
\end{figure*}

\paragraph{Qualitative Evaluation}
Figure \ref{fig:metric_failures} demonstrates how individual RS Score components fail in complementary ways, which our combined metric successfully addresses. MID assigns unreasonably low scores to valid captions due to distributional modeling limitations. DINO penalizes captions too harshly for minor visual inaccuracies that humans find acceptable. LLM embeddings give high scores to linguistically fluent but visually misaligned captions. In each case, the integrated RS Score compensates for these individual weaknesses, producing scores that better match human judgments by leveraging the strengths of other components.

\paragraph{Statistical Robustness Analysis}
To assess the stability of our correlation results, we performed 1000-run bootstrap analysis on Flickr8k. The results are presented in Table~\ref{tab:bootstrap_robustness}, which shows that our proposed metric exhibits excellent stability ($\sigma = 0.43\%$), confirming the statistical robustness of our findings. The Redemption Score consistently achieves the highest correlation ($58.42 \pm 0.43\%$) with the most stable performance across bootstrap samples.

\begin{table}[htbp]
\centering
\caption{Statistical Robustness Analysis: 1000-run Bootstrap Results on Flickr8k}
\label{tab:bootstrap_robustness}
\renewcommand{\arraystretch}{1.3}
\begin{tabular}{lccc}
\toprule
\textbf{Metric} & \textbf{Mean (\%)} & \textbf{Std Dev (\%)} & \textbf{95\% CI} \\
\midrule
\rowcolor{oursA} RS & 58.4 & 0.43 & [57.6, 59.2] \\
MID & 54.6 & 0.50 & [53.7, 55.6] \\
GTEScore & 53.9 & 0.48 & [53.0, 54.8] \\
CLIP & 51.1 & 0.48 & [50.2, 52.1] \\
DINO & 48.8 & 0.48 & [47.8, 49.7] \\
BERTScore & 33.6 & 0.58 & [32.6, 34.8] \\
\bottomrule
\end{tabular}
\end{table}

\paragraph{Results on Conceptual Captions.} Since human evaluations are unavailable for this dataset, we evaluate generalizability using the fixed parameters learned from Flickr8k. This tests whether RS captures transferable aspects of caption quality across different visual domains and annotation styles.

The results from this study is highlighted on \cref{tab:model_metrics_combined}. As noted previously on \cref{sec:captioning_models}, all 5 image captioning multimodal systems were fine tuned on the Conceptual Captions dataset thus leading to a higher Redemption Score.

\paragraph{Results on MS COCO. } A similar framework of evaluation was employed to assess Redemption Score's capacity of consistently portraying the model's capacity to caption images. Since the models were fine-tuned over the Conceptual Captions dataset and the captioning style and distribution of the Conceptual Captions and COCO are slightly different, the exact numbers were not replicated but the general trend of Qwen2-VL and the BLIP-2 leading the pact was observed in \cref{tab:model_metrics_combined}. Only slight changes in the score's trend was observed on GIT and ViT-GPT2 but this could be attributed towards the comparatively smaller models overfitting to the fine-tuned dataset. 

\subsection{\textcolor{black}{Additional Cross-Dataset Validation}}
\textcolor{black}{To evaluate whether Redemption Score generalizes beyond the datasets used during metric development, we additionally evaluate it on Pascal-50S and Nebula. Importantly, the fusion weights $(\alpha,\beta,\gamma,\lambda)=(0.15,0.35,0.50,0.80)$ are selected exclusively on Flickr8k and remain fixed for all experiments reported in this section. No dataset-specific tuning is performed on either Pascal-50S or Nebula. Consequently, these experiments evaluate cross-dataset hyperparameter transfer rather than adaptation to the target benchmarks.} \textcolor{black}{\paragraph{Pairwise human preference on Pascal-50S.}} \textcolor{black}{Table~\ref{tab:pascal50s_redemption} reports pairwise preference accuracy across human--human correct (HC), human--human incorrect (HI), human--machine (HM), and machine--machine (MM) comparisons. Redemption Score achieves a mean accuracy of 84.3, remaining within 0.9 percentage points of MID, which obtains 85.2. Although MID performs better on the HC and HM categories, Redemption Score achieves 99.8 on HI and 78.3 on MM, compared with 99.7 and 76.8 for MID, respectively.} \textcolor{black}{The improvement on the MM category is particularly relevant because this setting requires the metric to distinguish between two competing machine-generated captions rather than separating a machine-generated caption from a human reference. These results indicate that the proposed fusion remains competitive under pairwise human preference evaluation and can provide improved discrimination between closely competing generated descriptions. Importantly, this performance is obtained using the original Flickr8k-selected weights without any retuning on Pascal-50S.} \textcolor{black}{\paragraph{Correlation with human judgments on Nebula.}} \textcolor{black}{Table~\ref{tab:nebula_redemption} reports correlations between metric scores and human judgments on the Nebula test set. Redemption Score achieves Kendall $\tau_b=0.4996$, Kendall $\tau_c=0.4737$, and Spearman $\rho=0.6757$. These rank-based correlations are slightly lower than the corresponding MID results of $0.5211$, $0.4941$, and $0.6969$, respectively. In contrast, Redemption Score achieves a Pearson correlation of $0.8363$, exceeding the MID result of $0.8159$.} \textcolor{black}{This difference suggests that MID retains a modest advantage in pairwise ranking consistency, whereas Redemption Score exhibits stronger linear agreement with the magnitude of the human ratings. Thus, the proposed fusion does not uniformly improve every correlation measure, but it provides competitive rank-based agreement and stronger Pearson alignment without any Nebula-specific hyperparameter optimization.} \textcolor{black}{\paragraph{Cross-dataset robustness of the fusion weights.}} \textcolor{black}{Taken together, the Pascal-50S and Nebula results provide evidence that the weights selected on Flickr8k remain effective under different datasets, annotation protocols, and evaluation objectives. Redemption Score remains within 0.9 percentage points of MID on the Pascal-50S mean score, improves the MM category by 1.5 percentage points, and increases Pearson correlation on Nebula by 0.0204. These results indicate that the fixed fusion does not collapse when transferred beyond the development distribution.} \textcolor{black}{At the same time, the differences across Pascal-50S categories and Nebula correlation measures demonstrate that no single fixed fusion is uniformly optimal for every target distribution. We therefore interpret these experiments as evidence of competitive cross-dataset robustness rather than universal superiority over MID. Further evaluation on specialized visual domains and long-form multimodal reasoning responses remains an important direction for future work.}
\begin{table}[t]
\centering
\caption{\textcolor{black}{Pascal-50S pairwise preference accuracy. Redemption Score uses fixed Flickr8k-tuned weights $(\alpha,\beta,\gamma,\lambda)=(0.15,0.35,0.50,0.80)$.}}
\label{tab:pascal50s_redemption}
\begin{tabular}{lccccc}
\hline
\textcolor{black}{Method} & \textcolor{black}{HC} & \textcolor{black}{HI} & \textcolor{black}{HM} & \textcolor{black}{MM} & \textcolor{black}{Mean} \\
\hline
\textcolor{black}{Length} & \textcolor{black}{51.7} & \textcolor{black}{52.3} & \textcolor{black}{63.6} & \textcolor{black}{49.6} & \textcolor{black}{54.3} \\
\textcolor{black}{BLEU-4} & \textcolor{black}{60.4} & \textcolor{black}{90.6} & \textcolor{black}{84.9} & \textcolor{black}{54.7} & \textcolor{black}{72.6} \\
\textcolor{black}{SPICE} & \textcolor{black}{63.6} & \textcolor{black}{96.3} & \textcolor{black}{86.7} & \textcolor{black}{68.3} & \textcolor{black}{78.7} \\
\textcolor{black}{METEOR} & \textcolor{black}{63.8} & \textcolor{black}{97.7} & \textcolor{black}{93.7} & \textcolor{black}{65.4} & \textcolor{black}{80.1} \\
\textcolor{black}{ROUGE-L} & \textcolor{black}{63.7} & \textcolor{black}{95.3} & \textcolor{black}{92.3} & \textcolor{black}{61.2} & \textcolor{black}{78.1} \\
\textcolor{black}{CIDEr} & \textcolor{black}{65.1} & \textcolor{black}{98.1} & \textcolor{black}{90.5} & \textcolor{black}{64.8} & \textcolor{black}{79.6} \\
\textcolor{black}{BERT-S (RoBERTa-F)} & \textcolor{black}{65.4} & \textcolor{black}{96.2} & \textcolor{black}{93.3} & \textcolor{black}{61.4} & \textcolor{black}{79.1} \\
\textcolor{black}{TIGEr} & \textcolor{black}{56.0} & \textcolor{black}{99.8} & \textcolor{black}{92.8} & \textcolor{black}{74.2} & \textcolor{black}{80.7} \\
\textcolor{black}{ViLBERTScore-F} & \textcolor{black}{49.9} & \textcolor{black}{99.6} & \textcolor{black}{93.1} & \textcolor{black}{75.8} & \textcolor{black}{79.6} \\
\textcolor{black}{BERT-S++} & \textcolor{black}{65.4} & \textcolor{black}{98.1} & \textcolor{black}{96.4} & \textcolor{black}{60.3} & \textcolor{black}{80.1} \\
\textcolor{black}{CLIP-S} & \textcolor{black}{56.5} & \textcolor{black}{99.3} & \textcolor{black}{96.4} & \textcolor{black}{70.4} & \textcolor{black}{80.7} \\
\textcolor{black}{RefCLIP-S} & \textcolor{black}{64.5} & \textcolor{black}{99.6} & \textcolor{black}{95.4} & \textcolor{black}{72.8} & \textcolor{black}{83.1} \\
\textcolor{black}{MID} & \textcolor{black}{67.0} & \textcolor{black}{99.7} & \textcolor{black}{97.4} & \textcolor{black}{76.8} & \textcolor{black}{85.2} \\
\textcolor{black}{Redemption Score} & \textcolor{black}{64.1} & \textcolor{black}{99.8} & \textcolor{black}{95.0} & \textcolor{black}{78.3} & \textcolor{black}{84.3} \\
\hline
\end{tabular}
\end{table}

\begin{table}[t]
\centering
\caption{\textcolor{black}{Nebula test-set correlation with human scores. Redemption Score uses fixed Flickr8k-tuned weights $(\alpha,\beta,\gamma,\lambda)=(0.15,0.35,0.50,0.80)$.}}
\label{tab:nebula_redemption}
\begin{tabular}{lcccc}
\hline
\textcolor{black}{Method} & \textcolor{black}{Kendall $\tau_b$} & \textcolor{black}{Kendall $\tau_c$} & \textcolor{black}{Spearman} & \textcolor{black}{Pearson} \\
\hline
\textcolor{black}{DINO} & \textcolor{black}{0.4685} & \textcolor{black}{0.4442} & \textcolor{black}{0.6426} & \textcolor{black}{0.7367} \\
\textcolor{black}{GTE} & \textcolor{black}{0.5100} & \textcolor{black}{0.4836} & \textcolor{black}{0.6864} & \textcolor{black}{0.7809} \\
\textcolor{black}{MID} & \textcolor{black}{0.5211} & \textcolor{black}{0.4941} & \textcolor{black}{0.6969} & \textcolor{black}{0.8159} \\
\textcolor{black}{Redemption Score} & \textcolor{black}{0.4996} & \textcolor{black}{0.4737} & \textcolor{black}{0.6757} & \textcolor{black}{0.8363} \\
\hline
\end{tabular}
\end{table}

\subsection{Ablation Study}

\paragraph{Purely Multiplicative Strategy.}
To validate our hybrid aggregation approach in \cref{eq:redeem}, we test a purely multiplicative strategy without the additive component ($\lambda = 0$):
\begin{align}
 \hat{RS}(I,\hat{c};R) = MID^\alpha \cdot DINO_{sim}^\beta \cdot GTEScore^\gamma\\
 \alpha+\beta+\gamma=1; \alpha,\beta,\gamma \geq 0 \nonumber
\end{align}

\renewcommand{\arraystretch}{1.2}
\begin{tabular}{lccc}
\toprule
Best Weights ($\alpha, \beta, \gamma$) & Kendall's $\tau$ & Mean Score \\
\midrule
(0.150, 0.200, 0.650) & 0.5713 & 0.5141 \\
\bottomrule
\end{tabular}

\vspace{1em}  

\paragraph{Purely Additive Strategy.}
We also test a purely additive approach without the multiplicative component ($\lambda = 1$):
\begin{align}
 \hat{RS}(I,\hat{c};R) = \alpha \cdot MID + \beta \cdot DINO_{sim} + \gamma \cdot GTEScore\\
 \alpha+\beta+\gamma=1; \alpha,\beta,\gamma \geq 0 \nonumber
\end{align}

\renewcommand{\arraystretch}{1.2}
\begin{tabular}{lccc}
\toprule
Best Weights ($\alpha, \beta, \gamma$) & Kendall's $\tau$ & Mean Score \\
\midrule
(0.350, 0.250, 0.400) & 0.5826 & 0.5405 \\
\bottomrule
\end{tabular}

\vspace{1em}  

\paragraph{Comparative Analysis.}
Table~\ref{tab:ablation} presents a statistical comparison of the three approaches based on 
1000-run bootstrap analysis. The column \emph{Mean} reports the average $\tau$ across bootstrap runs, 
$\boldsymbol{\sigma}$ is the corresponding standard deviation, 
\emph{95\% CI} denotes the confidence interval, and \emph{Width} is the span of that interval. 
\emph{Imp.} indicates relative improvement over the baseline.

\begin{table}[h]
\centering
\footnotesize
\caption{Ablation Study: Robustness Comparison}
\label{tab:ablation}
\renewcommand{\arraystretch}{1.3}
\begin{tabular}{l@{\hskip 6pt}c@{\hskip 6pt}c@{\hskip 6pt}c@{\hskip 6pt}c@{\hskip 6pt}c}
\toprule
\textbf{Approach} & \textbf{Mean} & \textbf{$\boldsymbol{\sigma}$} & \textbf{95\% CI} & \textbf{Width} & \textbf{Imp.} \\
\midrule
\rowcolor{oursA}{Hybrid} & {0.5839} & {0.0043} & {[0.5750,\,0.5923]} & {0.0174} & {Base} \\
Additive & 0.5826 & 0.0046 & [0.5734,\,0.5915] & 0.0181 & -0.22\% \\
Multiplicative & 0.5713 & 0.0045 & [0.5624,\,0.5803] & 0.0179 & -2.21\% \\
\bottomrule
\end{tabular}
\end{table}

The hybrid method outperforms both additive and multiplicative approaches, achieving the highest mean $\tau$ (0.5839), the narrowest CI width (0.0174), and improvements of +0.22\% and +2.21\% respectively.

\paragraph{Metric Choice}
To ensure our metric selection is both principled and robust, we conducted an exhaustive ablation over all 20 possible three-metric combinations from MID, General Text Embeddings(GTE), DINO, BERT, LPIPS, and CLIP. For LPIPS, we use a normalized variant $LPIPS_{norm} = \tfrac{1}{1+LPIPS}$ to maintain comparability across scales. As shown in Table~\ref{tab:results_new}, our chosen combination (MID + LLM + DINO) achieves the highest correlation with human judgments ($\tau = 0.584$) while maintaining a low standard deviation, demonstrating both accuracy and stability. Importantly, most of the top-performing combinations draw from different buckets of evaluation aspects: MID provides distributional alignment, LLM embeddings and BERT capture semantic grounding, DINO and LPIPS measure visual fidelity, and CLIP captures cross-modal alignment. This pattern highlights that the strongest evaluators are not those relying on a single perspective, but those that integrate complementary dimensions of textual meaning, visual similarity, and multimodal consistency. By explicitly combining metrics across these buckets, our selected approach provides the most comprehensive and reliable evaluation signal.

\begin{table}[h]
\footnotesize
\centering
\caption{Ablation Results for evaluation of different metrics}
\label{tab:results_new}
\setlength{\tabcolsep}{4pt}
\renewcommand{\arraystretch}{1.3}
\begin{tabular}{@{}lcccc@{}}
\toprule
\textbf{Metric Combination} & \textbf{Weights ($\alpha,\beta,\gamma$)} & \textbf{$\lambda$} & \textbf{Kendall $\tau$} & \textbf{Std Dev} \\
\midrule
\rowcolor{oursA}{MID + GTEScore + DINO} & {0.15, 0.50, 0.35} & {0.80} & {0.584} & {0.106} \\
MID + GTEScore + LPIPS & 0.20, 0.50, 0.30 & 0.80 & 0.581 & 0.118 \\
MID + GTEScore + BERT & 0.15, 0.65, 0.20 & 0.60 & 0.578 & 0.125 \\
MID + GTEScore + CLIP & 0.15, 0.65, 0.20 & 0.00 & 0.574 & 0.386 \\
GTEScore + DINO + CLIP & 0.55, 0.25, 0.20 & 0.00 & 0.572 & 0.144 \\
GTEScore + DINO + LPIPS & 0.50, 0.30, 0.20 & 1.00 & 0.567 & 0.041 \\
GTEScore + LPIPS + CLIP & 0.65, 0.15, 0.20 & 0.00 & 0.567 & 0.140 \\
MID + DINO + CLIP & 0.15, 0.60, 0.25 & 0.00 & 0.567 & 0.402 \\
GTEScore + BERT + CLIP & 0.65, 0.15, 0.20 & 0.00 & 0.565 & 0.159 \\
MID + BERT + DINO & 0.40, 0.15, 0.45 & 0.70 & 0.564 & 0.196 \\
GTEScore + BERT + DINO & 0.50, 0.15, 0.35 & 1.00 & 0.563 & 0.047 \\
MID + BERT + CLIP & 0.15, 0.40, 0.45 & 0.00 & 0.563 & 0.907 \\
MID + LPIPS + CLIP & 0.15, 0.50, 0.35 & 0.00 & 0.562 & 0.556 \\
MID + DINO + LPIPS & 0.50, 0.30, 0.20 & 0.80 & 0.561 & 0.216 \\
MID + BERT + LPIPS & 0.55, 0.15, 0.30 & 0.70 & 0.556 & 0.237 \\
GTEScore + BERT + LPIPS & 0.60, 0.15, 0.25 & 1.00 & 0.549 & 0.043 \\
BERT + DINO + CLIP & 0.15, 0.60, 0.25 & 0.00 & 0.545 & 0.173 \\
DINO + LPIPS + CLIP & 0.60, 0.15, 0.25 & 0.00 & 0.543 & 0.155 \\
BERT + LPIPS + CLIP & 0.25, 0.55, 0.20 & 0.10 & 0.528 & 0.203 \\
BERT + DINO + LPIPS & 0.15, 0.65, 0.20 & 1.00 & 0.498 & 0.038 \\
\bottomrule
\end{tabular}
\end{table}

\subsection{\textcolor{black}{Efficiency and Runtime Analysis}}
\label{sec:efficiency_runtime}

\textcolor{black}{
Redemption Score introduces additional computational cost because its
DINO-based component requires generating visual proxies with Stable
Diffusion~3 (SD3). To quantify this overhead, we compare its wall-clock
runtime with conventional captioning metrics and a VLM-as-a-Judge
baseline on Nebula. Table~\ref{tab:runtime_overhead} reports both total
runtime and average latency per sample. Traditional metrics and the
VLM-as-a-Judge baseline are evaluated on subsets of 512 and 20 samples,
respectively, whereas Redemption Score is evaluated using the complete
Nebula test set of 3,298 samples. Because the evaluated sample counts
differ, average time per sample provides the most direct comparison.
All timings include model initialization and input/output operations as
recorded by the corresponding evaluation pipelines.
}
\begin{table*}[!t]
\centering
\caption{\textcolor{black}{
Runtime and execution comparison on Nebula. Traditional metrics are
timed on 512 samples, Qwen3.6-27B VLM-as-a-Judge on 20 samples, and
Redemption Score on the complete Nebula test split of 3,298 samples.
Redemption Score timings are obtained from the full evaluation logs.
Wall-clock time includes model initialization and I/O where applicable.
}}
\label{tab:runtime_overhead}

{\color{black}
\small
\setlength{\tabcolsep}{5pt}
\renewcommand{\arraystretch}{1.12}

\begin{tabular}{@{}p{4.4cm}ccc p{5.2cm}@{}}
\hline
\textbf{Method}
&
\textbf{Samples}
&
\textbf{Total Time}
&
\shortstack{\textbf{Avg. Time}\\\textbf{/ Sample}}
&
\textbf{Execution Characteristics}
\\
\hline

BLEU-4
& 512
& 0.15 s
& 0.0003 s
& Lightweight lexical metric; no large neural model required.
\\

ROUGE-L
& 512
& 0.77 s
& 0.0015 s
& Lightweight lexical metric; no large neural model required.
\\

CIDEr
& 512
& 0.68 s
& 0.0013 s
& Lightweight consensus-based metric; no large generative model required.
\\

BERTScore
& 512
& 19.35 s
& 0.0378 s
& Embedding-based inference using a comparatively small language encoder.
\\

Qwen3.6-27B VLM-as-a-Judge
& 20
& 84.34 s
& 4.2168 s
& The VLM required quantization to 4 bits to fit the images and the captions alongside the KV cache into the Nvidia A40-48Q GPU.
\\

Qwen2.5-vl-72b-instruct VLM-as-a-Judge& --
&-- 
&-- 
& Does not fit on the GPU
\\

\hline

RS scoring with cached proxy images
& 3,298
& 9.05 min
& 0.1646 s
& DINO and text-component scoring only; proxy images are reused from disk.
\\

SD3 candidate-proxy generation
& 3,298
& 6.79 h
& 7.4121 s
& Executed on a single GPU; each sample is independent and can be
distributed across multiple GPUs.
\\

SD3 reference-proxy generation
& 3,298
& 6.79 h
& 7.4136 s
& One-time preprocessing cost for a fixed benchmark; generated reference
proxies can be reused across model evaluations.
\\

RS end-to-end with references cached
& 3,298
& 6.94 h
& 7.5767 s
& Requires candidate-proxy generation, while benchmark reference proxies
are generated once and reused.
\\

RS end-to-end, fully uncached
& 3,298
& 13.73 h
& 14.9903 s
& Includes both candidate- and reference-proxy generation together with
the final Redemption Score computation.
\\

\hline
\end{tabular}
}
\end{table*}

\textcolor{black}{
As expected, lexical and embedding-based metrics remain substantially
faster than Redemption Score because they do not require image
generation. Once candidate and reference proxy images have been cached,
Redemption Score requires 0.1646 seconds per sample, compared with
4.2168 seconds per sample for the Qwen3.6-27B VLM-as-a-Judge baseline.
Thus, cached Redemption Score scoring is approximately
$25.6\times$ faster than the evaluated VLM judge, although it remains
slower than conventional embedding-only and lexical metrics.
}

\textcolor{black}{
The primary computational bottleneck is SD3 proxy generation rather
than the subsequent Redemption Score computation. Generating one set of
candidate proxies requires approximately 7.4121 seconds per sample,
whereas generating both candidate and reference proxies increases the
fully uncached end-to-end latency to 14.9903 seconds per sample. Under
the fully uncached setting, Redemption Score is approximately
$3.55\times$ slower than the evaluated VLM-as-a-Judge baseline.
Therefore, the quantitative results confirm the reviewer's concern that
the proposed metric has a meaningful one-time generative overhead.
}

\textcolor{black}{
For evaluation on a fixed benchmark, however, reference captions do not
change and their proxy images can be generated once and reused. In this
more representative setting, evaluating a new collection of candidate
outputs requires candidate-proxy generation followed by cached
Redemption Score computation, resulting in approximately 7.5767 seconds
per sample. This remains approximately $1.80\times$ slower than the
measured VLM-as-a-Judge baseline, but reduces the cost relative to fully
uncached Redemption Score by nearly half. Proxy generation can also be
distributed independently across samples and parallelized across
multiple GPUs.
}

\textcolor{black}{
These measurements clarify that Redemption Score is intended primarily
for offline benchmark evaluation, model auditing, and detailed
hallucination analysis rather than latency-sensitive online evaluation.
Its additional cost provides an explicit generative grounding signal,
but this benefit comes with a clear efficiency trade-off. Future work
could reduce this overhead through precomputed reference-proxy banks,
batched and parallel generation, fewer diffusion steps, or lighter
proxy-generation models. Because the VLM-as-a-Judge timing is measured
on only 20 samples, we treat that comparison as an indicative latency
estimate rather than a definitive throughput benchmark.
}

\section{Limitations, Conclusion and Future Works}
\paragraph{Limitations. }
\textcolor{black}{
 Because RS is built on pretrained foundation models, it can
inherit their biases and failure modes. The triangulation strategy should therefore not be interpreted as eliminating such biases. Instead, it attenuates the effect of individual component failures by combining complementary signals: MID captures distributional image-text alignment, DINO-based similarity captures cycle-based visual consistency, and GTEScore captures linguistic-semantic agreement. As shown in Fig.~\ref{fig:metric_failures}, individual components can
fail in different ways, while the fused RS score often compensates by relying on the remaining signals. However, if multiple components share a systematic bias, or if the T2I generation step introduces consistent artifacts, that bias
may still propagate into RS. We therefore view RS as a bias-diluting and diagnostic framework rather than a bias-free metric.
}
\paragraph{Conclusion. } We introduce Redemption Score, a multi-modal evaluation framework that achieves 58.42\% Kendall-$\tau$ correlation with human judgments on Flickr8k, outperforming existing methods. RS addresses individual metric limitations through calibrated parameter optimization and demonstrates robust generalization across Conceptual Captions and MS-COCO datasets without requiring task-specific training. The strength of this work lies in its interpretable nature and decomposable structure, where component-wise scoring reveals specific failure modes such as low DINO scores indicating visual grounding issues, low GTE scores revealing linguistic inadequacy, and low MID scores suggesting distributional anomalies. Patterns with variable performance like high GTE with low DINO expose hallucination problems, allowing targeted interventions and architectural changes. Researchers can use component scores to systematically diagnose models and guide improvements in visual grounding, linguistic quality, or distributional alignment. While the framework is currently constrained by the computational overhead associated with cycle-consistent image generation via Stable Diffusion 3, the resulting gains in evaluative precision offer a much more nuanced signal. This methodology allows practitioners to move beyond scalar rankings toward a systematic analysis of visual-semantic alignment.

\paragraph{Future Works. } Key extensions include: (i) multilingual support across diverse linguistic and cultural contexts, requiring adaptation of text encoders and cross-cultural validation to ensure evaluation reliability across different annotation traditions and captioning conventions, (ii) computational efficiency through model distillation and lightweight alternatives to full image generation, significantly reducing evaluation overhead while preserving the framework's diagnostic capabilities for practical deployment in resource-constrained settings, and (iii) temporal grounding by extending from static image-text to video-text evaluation, incorporating temporal coherence assessment and action grounding mechanisms to evaluate whether captions accurately describe event sequences and dynamic content in video data.


\bibliographystyle{IEEEtran}
\bibliography{main}

\end{document}